\documentclass{article}

\usepackage{microtype}
\usepackage{graphicx}
\usepackage{subcaption}
\usepackage{booktabs}
\usepackage{blindtext}
\usepackage{amsmath}
\usepackage{dsfont}
\usepackage{bbm}
\usepackage{graphicx}
\usepackage{bbm}
\usepackage{scrextend}
\usepackage{amsmath,amsfonts}

\usepackage{makecell}

\usepackage{hyperref}

\newcommand*{\Scale}[2][4]{\scalebox{#1}{$#2$}}%
\DeclareMathOperator*{\argmin}{arg\,min}
\DeclareMathOperator*{\argmax}{arg\,max}

\usepackage[accepted]{icml2018}

\icmltitlerunning{Learning to rank for censored survival data}

\begin{document}

\twocolumn[
\icmltitle{Learning to rank for censored survival data}

\icmlsetsymbol{equal}{*}

\begin{icmlauthorlist}
\icmlauthor{Margaux Luck}{equal,mila}
\icmlauthor{Tristan Sylvain}{equal,mila}
\icmlauthor{Joseph Paul Cohen}{mila}
\icmlauthor{H\'elo\"ise Cardinal}{chum}
\icmlauthor{Andrea Lodi}{poly}
\icmlauthor{Yoshua Bengio}{mila}
\end{icmlauthorlist}

\icmlaffiliation{mila}{Montreal Institute for Learning Algorithms, Universit\'e de Montr\'eal,  Montr\'eal, Canada}
\icmlaffiliation{poly}{Canada Excellence Research Chair, \'Ecole Polytechnique de Montr\'eal,  Montr\'eal, Canada}
\icmlaffiliation{chum}{Center of Research and Department of Medicine, Nephrology, Centre Hospitalier de l'Universit\'e de Montr\'eal,  Montr\'eal, Canada}

\icmlcorrespondingauthor{Margaux Luck, Tristan Sylvain}{\{luckmarg, sylvaint\}@iro.umontreal.ca}

\icmlkeywords{Deep Learning, Survival analysis, Ranking, Wasserstein metric}

\vskip 0.3in
]

\printAffiliationsAndNotice{\icmlEqualContribution}

\begin{abstract}
Survival analysis is a type of semi-supervised ranking task where the target output (the survival time) is often right-censored. Utilizing this information is a challenge because it is not obvious how to correctly incorporate these censored examples into a model. We study how three categories of loss functions, namely partial likelihood methods, rank methods, and our classification method based on a Wasserstein metric (WM) and the non-parametric Kaplan Meier estimate of the probability density to impute the labels of censored examples, can take advantage of this information. The proposed method allows us to have a model that predict the probability distribution of an event. If a clinician had access to the detailed probability of an event over time this would help in treatment planning. For example, determining if the risk of kidney graft rejection is constant or peaked after some time. Also, we demonstrate that this approach directly optimizes the expected C-index which is the most common evaluation metric for ranking survival models.
\end{abstract}

\section{Introduction}
Survival analysis, also known as time-to-event analysis aims to predict the first time of the occurrence of a stochastic event, conditioned on a set of features. An example in the case of medical data is the time of death or a graft failure after an operation. In cases where the time of event for many samples is missing because the event wasn't observed, this can be framed as a particular type of semi-supervised learning where part of the target values are referred to as right-censored. Formally we can say that for some examples we do not have the time of event $T$, but rather a time $T_0$ (censoring time) such that we know $T > T_0$. The classical approach to survival analysis is the Cox proportional hazards model \cite{cox1972regression} that takes into account censored samples. Ranking approaches \cite{steck2008ranking} are also a way to take these censored samples into account by incorporating them into the training using pairwise ranking loss where although the exact time of event is not known the pairwise relationship with respect to a censoring date is known for event occurring before the censored event. We would like to predict the probability distribution of an event as it will help in treatment planning. For example, determining if the risk of kidney graft rejection is constant or peaked after some time.

In this study, we propose to use the Wasserstein metric to have a model predict the probability distribution of the event time. This approach not only provides an interpretable prediction but allows us to impute the distribution of censored samples given global survival statistics with the non-parametric Kaplan Meier estimate. Our intuition is that training with the KM estimate provides a richer signal during training than a rank loss would provide. Also, we find that this approach directly optimizes the C-index \cite{harrell1982evaluating} which is the most common evaluation metric for ranking survival models. We compared our proposed loss with a set of common ranking-specific losses on several reference survival datasets.

\section{Survival data}
In what follows, we will use the following notations. 
Let $\mathbf{x}^{(i)}$ be the feature vector of the $i$-th example and let $\mathbf{y}^{(i)}_t$ take value 1 if event $i$ happened at time $t$ and 0 otherwise. Moreover, let $\hat{\mathbf{y}}^{(i)}_t$ be the estimated probability of event $i$ happening at time $t$ and let $t^{(i)}$ be the (scalar) actual time of event $i$. We denote by $\mathbf{z}^{(i)}_t$ and $\hat{\mathbf{z}}^{(i)}_t$ the true and estimated cumulative probability distribution of $y$. Namely, $\mathbf{z}^{(i)}_{t_0} = \sum_{t < t_0} \mathbf{y}^{(i)}_t$. Finally, let $c^{(i)}$ be 1 if example $i$ is observed (non-censored) and 0 otherwise.

\subsection{Ties and censored data}
Survival datasets describe medical events that can have a low temporal resolution (time scale) causing ties between patients. A given \emph{unique time} (at a given resolution, e.g., one day) can correspond to multiple events. Such events are \emph{tied} and that would imply that more precise predictions are not relevant. However, they must be given special attention in constructing loss functions.

As mentioned earlier, another characteristic of survival data is that they are right-censored. We can still use these examples by only comparing with patients that had an event before the date of censorship or by imputing the event time based on statistics over the data.

\subsection{Metric of evaluation}
The concordance index or C-index \cite{harrell1982evaluating} is the standard evaluation metric for survival data. It corresponds to the normalized Kendall tau metric between the true and predicted distribution \cite{kendall1938new}. It can be seen as a generalization of the Area Under the Receiver Operating Characteristic Curve (AUROC) that can handle right-censored data \cite{steck2008ranking}.

We define an \emph{acceptable} pair as one for which we are sure the first event occurs before the second. These are the pairs for which the first element is non-censored, and for which the censoring or event time of the second element is strictly greater than the first. Let $\mathcal{A}$ be the set of acceptable pairs.
Then, the C-index to be maximized can be written as:%
$$
\frac{1}{|\mathcal{A}|} \sum_{\scriptscriptstyle (\mathbf{x}^{(i)}, \mathbf{x}^{(j)}) \in \mathcal{A}} \Scale[0.8]{ \mathds{1}\Big(f(\mathbf{x}^{(i)}) < f(\mathbf{x}^{(j)})\Big) + \frac{1}{2}\mathds{1}\Big(f(\mathbf{x}^{(i)}) = f(\mathbf{x}^{(j)})\Big)}.
$$
\vspace{-10pt}

\section{Loss functions for censored data}
In this section, we present loss functions in the context of survival prediction for censored data. We divide these loss functions into three categories: partial likelihood methods, rank methods, and our classification method based on a Wasserstein metric (WM).

\subsection{Cox Model}
Cox introduced a general conditional log-likelihood to fit survival models, in which the probability of observations is maximized \cite{cox1972regression}. It was demonstrated by \cite{steck2008ranking} that maximizing the Cox's partial likelihood is approximately equivalent to maximizing the C-index. We present the general formula, with a real-valued score prediction function $f_\theta$ estimating the probability of the event at a particular time, given input features $\mathbf{x}^{(i)}$. Denoting the predicted score $f_\mathbf{\theta}(\mathbf{x}^{(i)})$ the loss is:
\begin{equation*}
\ell (\theta) = \sum_{i:c^{(i)} = 1} \Big(\log f_\theta(\mathbf{x}^{(i)}) - \log \sum_{j : t^{(j)} \geq t^{(i)}} f_\theta(\mathbf{x}^{(j)}) \Big).
\end{equation*}%
We also consider a variant of this loss, Efron's approximation \cite{efron1977efficiency} that commonly improves performance when there are many tied event times.

In our experiments, the Cox variant refers to a multi-layer perceptron (MLP) $f_\theta$ trained with the normal Cox loss or with Efron's approximation loss, as in \cite{katzman2016deep, luck2017deep}.

\subsection{Ranking losses}
Many methods attempt to directly predict the rank of the different examples. This is done by learning the following objective:%
$$
\argmax_{\mathbf{\theta}} \frac{1}{|\mathcal{A}|} \sum_{(\mathbf{x}^{(i)}, \mathbf{x}^{(j)}) \in \mathcal{A}} \phi(f_{\mathbf{\theta}}(\mathbf{x}^{(i)}) - f_{\mathbf{\theta}}(\mathbf{x}^{(j)}))
$$%
where $\phi(z)$ is a function that relaxes the non-differentiable $\mathds{1}$ of the C-index~\cite{steck2008ranking}. 
We evaluated the functions used in~\cite{steck2008ranking}, Ranking SVM ~\cite{herbrich2000large}, Rankboost~\cite{freund2003efficient} and RankNet~\cite{burges2005learning}. These functions have been shown in~\cite{kalbfleisch1978non} to correspond to lower bounds on the C-index. We use $\sigma$ to denote the Sigmoid function $z \rightarrow \frac{1}{1+\exp(-z)}$.

\subsection{Wasserstein metric}
While there have to our knowledge been no previous attempts to use the Wasserstein metric on survival data or ranking problems,~\cite{frogner2015learning} used a Wasserstein loss for image classification and tag prediction.~\cite{hou2016squared} and ~\cite{beckham2017unimodal} apply a Wasserstein metric for the more restrictive case of ordinal classification. Recently,~\cite{mena2018learning} used the Sinkhorn algorithm, which is commonly used in optimal transport applications, as an analogy to the Softmax for permutations.

The WM is the minimum cost to transport the mass from one probability distribution to another. In the case of distributions of discrete supports (histograms of class probabilities), this is computed by moving probability mass from one class to another, according to the ground distance matrix specifying the cost to transport probability mass to and from different classes. Thus, the WM takes advantage of knowledge of the structure of the space of values considered, e.g., the 1-dimensional real-valued time axis, so that some errors (e.g. between neighboring events) are appropriately penalized less than others.

The WM is particularly adapted to a survival context. We denote $p_r$ the true data distribution, and $p_\theta$ the distribution estimated by the model. We write $\Pi$ the set of joint distributions $p(\cdot, \cdot)$ with left and right marginals $p_\theta$ and $p_r$ respectively. Given an example $\mathbf{x}$ and corresponding real time of event $T$, we can write:%
$$
W(p_\theta, p_r) = \inf_{p(\cdot, \cdot | \mathbf{x}) \in \Pi} \mathbb{E}_{T_1, T_2 \sim p(\cdot, \cdot | \mathbf{x})} \big[d(T_1, T_2) \big]
$$%
As $p_r$ is a Dirac, we have that:%
$$
\mathbb{E}_{T_1, T_2 \sim p(\cdot, \cdot | \mathbf{x})} \big[d(T_1, T_2) \big] = \mathbb{E}_{T_1 \sim p(\cdot, T | \mathbf{x})} \big[d(T_1, T) \big]
$$%
In all that follows, $d(T_1, T_2)$ is chosen to be proportional to the number of train set elements having events between $T_1$ and $T_2$. The term is therefore $\mathbb{E}_{T \sim p_\theta(\cdot, T | \mathbf{x})} \big[1-\text{Cindex} \big]$.

\subsubsection{Use as a learning objective}
\cite{levina2001earth} notes that under certain conditions satisfied in the case of ordinal classification, the WM takes the following expression:%
$$
\text{WM}(p, q) = \Big(\frac{1}{T}\Big)^{1/l} || CDF(p) - CDF(q) ||_l,
$$
where $T$ is the size of the Softmax layer and $CDF(.)$ is a function that returns the cumulative density function of its input density. Here, $p$ and $q$ are two probability distributions with discrete supports. We use $l=1.5$ in our experiments. We write $f_\theta (\mathbf{x}^{(i)}) = \hat{\mathbf{z}}^{(i)}_\theta$ to highlight the dependency on $\theta$. The objective can be written as:%
$$
\argmin_\theta \frac{1}{T}\sum_i  ||\hat{\mathbf{z}}^{(i)}_\theta - \mathbf{z}^{(i)} ||^l.
$$
\subsubsection{Imputing missing values for classification}
\begin{figure}[h]
\centering
\includegraphics[width=0.4\textwidth]{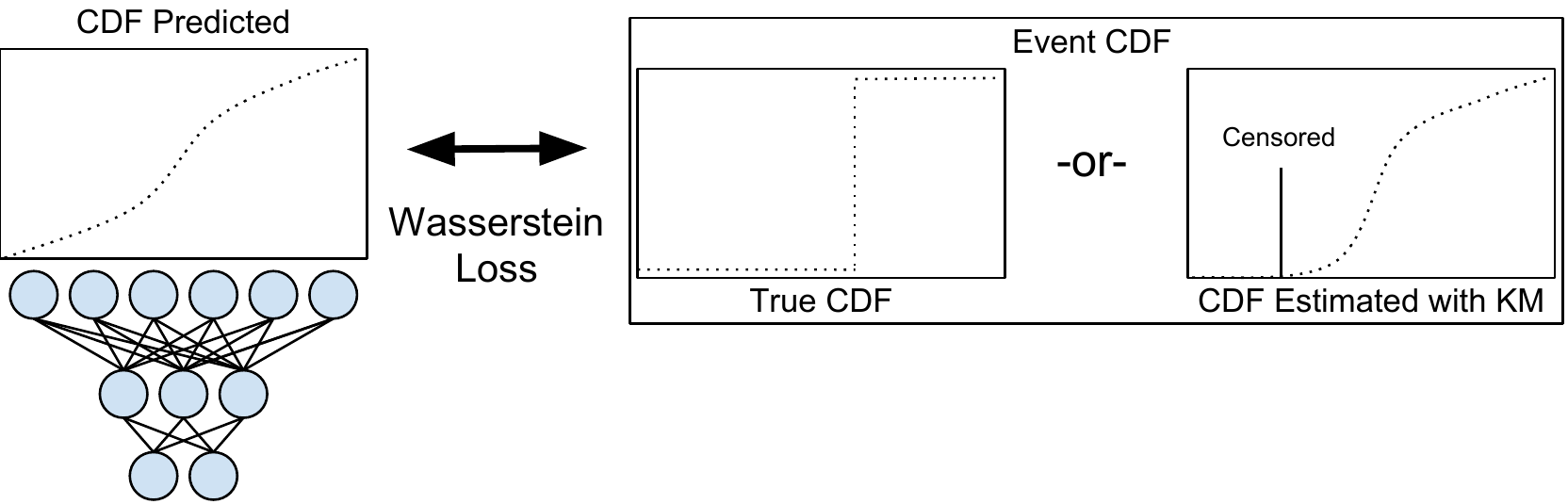}
\caption{An overview of the proposed distribution matching loss. In the case that a sample is censored the KM estimate is used to impute the probability that should be assigned for that event.}
\label{model}
\end{figure}

In order to allow the WM objective to lead to good training, we have imputed the CDF of the censored data with $1. - KM$, where $KM$ is the Kaplan-Meier non parametric estimate of the survival distribution function computed on the training set (see Figure~\ref{model}). With the KM estimator, the survival distribution function $S(t)$ is estimated as a step function, where the value at time $t_{i}$ is calculated as follows:
\begin{align*}
\hat{S}(t_{i}) = \hat{S}(t_{i-1})(1-d_{i}/n_{i}),
\end{align*}
with $d_{i}$ denoting the number of events at $t_{i}$ and $n_{i}$ the number of patients alive just before $t_{i}$. 

\section{Experiments}
\subsection{Datasets}
We assess the presented models on a variety of publicly available datasets. The characteristics of these datasets are summarized in Table \ref{tab:datasets}.

\begin{table}[!ht]
\resizebox{\columnwidth}{!}{%
\centering
\begin{tabular}{  l r r r r r  } 
\toprule
Datasets           & \thead{Nb.\\samples} & \thead{Nb. ($\%$) \\censored} & \thead{Nb. ($\%$) \\unique times } & \thead{Nb. \\features }\\ 
\midrule
SUPPORT2           & 9105        &  2904 (32.2)        & 1724 (19.1)              & 98        \\ 
AIDS3              & 3985        & 2223 (55.8)         & 1506 (37.8)              & 19        \\ 
COLON              & 929         & 477 (51.3)          & 780 (84.0)                & 48        \\ 

\bottomrule
\end{tabular}}
\caption{Characteristics of the datasets used in our evaluation. The datasets have different numbers of samples, percentage of censored, and tied patients. The features are typically continuous or discrete clinical attributes.}
\label{tab:datasets}
\end{table}
\vspace*{-40pt}
\textbf{SUPPORT2}\footnote{available at \protect\url{http://biostat.mc.vanderbilt.edu/wiki/Main/DataSets}} records the survival time for patients of the SUPPORT study.

\textbf{AIDS3}\footnote{available at \url{https://vincentarelbundock.github.io/Rdatasets/datasets.html}\label{first_footnote}} corresponds to the Australian AIDS Survival Data.

\textbf{COLON}\footref{first_footnote} consists of data from the first successful trials of adjuvant chemotherapy for colon cancer. We considered death as a target event for our study.

\subsection{Data pre-processing}
We used a one-hot encoding for categorical features, and unit scaling for continuous features. For features with missing values, we added an indicator function for the absence of a value. 

We performed 5 fold cross validation and kept 20\% of the train set as a validation set. The prediction performance was reported as mean $\pm$ standard error of the C-index over the 5 folds. Early stopping was performed on the validation C-index.

We used a multi-layer perceptron (3 layers with 100 units each) with ReLU activation functions where applicable, and used Dropout~\cite{hinton2012improving}, Batch Normalization~\cite{ioffe2015batch} and L2 regularization on the weights. We used the Adam optimizer. For the ranking and log-likelihood methods the output was a single unit with a linear activation function. For the methods requiring a prediction of output times, we used a Softmax function. Our code was written in PyTorch~\cite{paszke2017automatic}

We perform a grid-search for each split independently for the L2 regularization coefficient on the weight and the learning rate. We add a small constant (1 for Support2 and Aids3, 10 for Colon) to the distance between bins before normalizing. For colon we used a bin size of 2 days, and 1 day for the other two datasets.

\subsection{Comparison of different ranking methods}

We study the impact of the different loss functions in Table \ref{tab:results}. We study how the standard Cox model performs in comparison to ranking and classification losses.

\begin{table}[!ht]
\resizebox{\columnwidth}{!}{%
\centering
\begin{tabular}{ l l l l l l }
\toprule
Loss Type& Variant & SUPPORT2 & AIDS3 & COLON\\
\midrule 
 Partial likelihood & Cox & 84.90$\pm$0.63  & 54.84$\pm$0.82 & \textbf{64.66}$\pm$\textbf{0.44}\\
 Partial likelihood & Cox Efron's & 84.91$\pm$0.60 & 54.03$\pm$1.21 & 63.08$\pm$0.93 \\
 Ranking & $\sigma(z)$ & \textbf{85.53}$\pm$\textbf{0.56} & 55.35$\pm$1.19 & 64.22$\pm$0.61\\
 Ranking & Log-sigmoid & 85.44$\pm$0.57 & 55.28$\pm$1.29 & 63.36$\pm$0.52 \\
 Ranking & $(z-1)_+$ & 84.96$\pm$0.56 & 55.41$\pm$1.20 & 63.98$\pm$1.12 \\
 Ranking & $1 - \exp(-z)$ & 85.35$\pm$0.58 & 55.73$\pm$0.93 & 61.96$\pm$0.91 \\
 Classification & WM (ours) & 85.33$\pm$0.52 & \textbf{56.03}$\pm$\textbf{1.01} & 64.32$\pm$0.39 \\
\bottomrule
\end{tabular}}
\caption{Performance scores of the different methods. The table reports the C-index mean $\pm$ standard error over the 5 fold. For each dataset, the best model in terms of mean score is highlighted in bold. We draw the readers attention to the classification losses which are among the losses that give the best results.}
\label{tab:results}
\end{table}

\subsection{Impact of using censored data}
The purpose of this section is to explore how censoring is informative and demonstrate that we should not just ignore/process away censoring.
We compare three methods to account for censored data. We first completely removed censored examples from the training set (no censored data). We also considered the time of censoring to correspond to an actual event occurrence (transforming each example censored at time $t$ into the same example with an event occurring at time $t$) (death at censoring). Finally, we also listed results for the standard approach (with censored data). In the case of WM, the censored times are imputed with the ($1 - KM$) curve.

We run this experiment on the SUPPORT2 dataset for the three best methods of each category as it is the largest public dataset we have : Cox Efron's, $\sigma(z)$ and our methods WM. The results are presented in Table~\ref{table_censored}.

\begin{table}[h!]
\begin{center}
\resizebox{\columnwidth}{!}{%
\begin{tabular}{@{}cccc@{}}
\toprule
Method & WM & Ranking & Cox \\
\midrule
 No censored data & 83.31$\pm$0.51 & 83.40$\pm$0.52 &  82.34$\pm$0.49 \\ 
 Death at censoring & 82.34$\pm$0.58 & 81.97$\pm$0.67 & 80.67$\pm$0.55 \\
 With censored data & \textbf{85.33}$\pm$\textbf{0.52} & \textbf{85.53}$\pm$\textbf{0.56} & \textbf{84.91}$\pm$\textbf{0.60} \\
\bottomrule
\end{tabular}}
\end{center}
\caption{We explore how the three categories of methods are impacted by adding censored data. The table reports the C-index mean $\pm$ standard error over the 5 fold. For "Death at censoring", we set the death event as the censored time. It is clear that censored data contains information that we can use to make better predictions.}
\label{table_censored}
\end{table}

\subsection{Exploring the impact of censored data}
\begin{figure}[h]
\centering
\includegraphics[width=0.5\textwidth]{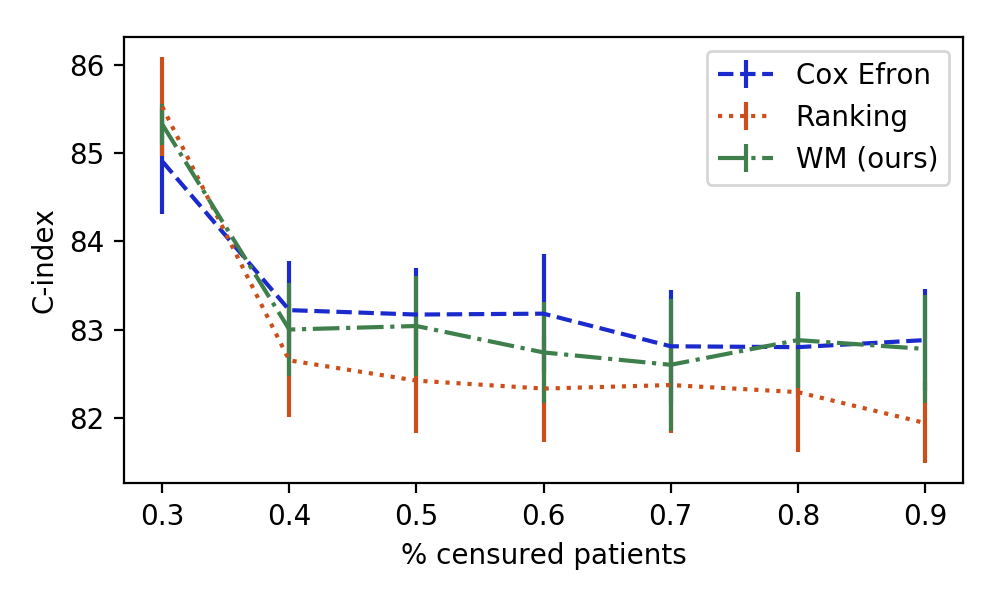}
\caption{Here we study how the composition of censored and uncensored patients during training impacts the C-index mean $\pm$ standard error over the 5 fold in the SUPPORT2 dataset. The validation and test sets are fixed and the training set has censored patients introduced by marking patients as censored at random. The plot starts at 30\% because the dataset has that many censored patients by default. We find that the WM classification loss is robust to the introduction of censored data.}
\label{varycensored}
\end{figure}

In order to determine how much of an improvement we can obtain from incorporating censored data we can vary the composition of samples that are censored in the training data, while keeping the validation and test sets the same. In Figure \ref{varycensored} we show the evolution of the C-index with different percentages of censoring of the training set in the SUPPORT2 dataset.

\section{Conclusion}
We proposed a new method for learning to rank survival data. Experiments on the different datasets show that our models trained with the WM loss gives accurate predictions compared to the more classical losses of the Cox model and ranking loss functions, which directly approximate a lower bound of the C-index. While not always state of the art, our method is always among the best results for each dataset.

We also find that this approach allows the method to tolerate a high percentage of censored samples and continue to predict well given results consistently in the same range of the best methods. Also, we demonstrate that our method can be seen as directly optimizing the expected C-index which is the most common evaluation metric for ranking survival models. Moreover, our results demonstrate that imputing the values with the KM curve for the missing times in a classification framework can increase the resulting C-index.

\section*{Acknowledgements}
We thank Christopher Pal and Christopher Beckham for their input on the project. This work is partially funded by a grant from the U.S. National Science Foundation Graduate Research Fellowship Program (grant number: DGE-1356104) and the Institut de valorisation des donnees (IVADO). This work utilized the supercomputing facilities managed by the Montreal Institute for Learning Algorithms, NSERC, Compute Canada, and Calcul Quebec.

\newpage
\bibliography{example_paper}
\bibliographystyle{icml2018-nopagenum}

\end{document}